%% file: main.tex
\documentclass[runningheads]{llncs}
\usepackage{graphicx}
\usepackage{tikz}
\usepackage{rotating}
\usepackage{comment}
\usepackage{amsmath,amssymb} % define this before the line numbering.
\usepackage{color}
\usepackage{subcaption}
\usepackage[accsupp]{axessibility}  % Improves PDF readability for those with disabilities.
\usepackage{breakcites}
\usepackage{adjustbox}
\usepackage{float}
\usepackage{longtable}

\usepackage{tabularx}
\usepackage[]{algorithm2e}
\usepackage{multirow, makecell} %For tables

\newcommand{\tikzxmark}{%
\tikz[scale=0.23] {
    \draw[line width=0.7,line cap=round] (0,0) to [bend left=6] (1,1);
    \draw[line width=0.7,line cap=round] (0.2,0.95) to [bend right=3] (0.8,0.05);
}}
\newcommand{\tikzcmark}{%
\tikz[scale=0.23] {
    \draw[line width=0.7,line cap=round] (0.25,0) to [bend left=10] (1,1);
    \draw[line width=0.8,line cap=round] (0,0.35) to [bend right=1] (0.23,0);
}}

\usepackage{hyperref}
\hypersetup{
    colorlinks=true,
    linkcolor=blue,
    filecolor=blue,      
    urlcolor=blue,
    citecolor=blue
}

\usepackage{booktabs}
\usepackage{xspace}

\newcommand\cifar{\texttt{CIFAR10}\xspace}
\usepackage{bm}

\begin{document}
% \linenumbers
\pagestyle{headings}
\mainmatter

\title{Flexible Channel Dimensions for Differentiable Architecture Search} % Replace with your title

%\begin{comment}
\titlerunning{} 
\authorrunning{A. C. Y\"uz\"ug\"uler et. al} 
\author{Ahmet Caner Y\"uz\"ug\"uler \and Nikolaos Dimitriadis \and Pascal Frossard}
\institute{EPFL  \\ 
\email{\{ahmet.yuzuguler,nikolaos.dimitriadis,pascal.frossard\}@epfl.ch}}
%\end{comment}
%******************

\begin{comment}
\titlerunning{Abbreviated paper title}
\author{First Author\inst{1}\orcidID{0000-1111-2222-3333} \and
Second Author\inst{2,3}\orcidID{1111-2222-3333-4444} \and
Third Author\inst{3}\orcidID{2222--3333-4444-5555}}
%
\authorrunning{F. Author et al.}
\institute{Princeton University, Princeton NJ 08544, USA \and
Springer Heidelberg, Tiergartenstr. 17, 69121 Heidelberg, Germany
\email{lncs@springer.com}\\
\url{http://www.springer.com/gp/computer-science/lncs} \and
ABC Institute, Rupert-Karls-University Heidelberg, Heidelberg, Germany\\
\email{\{abc,lncs\}@uni-heidelberg.de}}
\end{comment}

%******************
\maketitle

\begin{abstract}

% 150-words maximum

Finding optimal channel dimensions (i.e., the number of filters in DNN layers) is essential to design DNNs that perform well under computational resource constraints. 
Recent work in neural architecture search aims at automating the optimization of the DNN model implementation. 
However, existing neural architecture search methods for channel dimensions rely on fixed search spaces, which prevents achieving an efficient and fully automated solution.
In this work, we propose a novel differentiable neural architecture search method with an efficient dynamic channel allocation algorithm to enable a flexible search space for channel dimensions. 
We show that the proposed framework is able to find DNN architectures that are equivalent to previous methods in task accuracy and inference latency for the CIFAR-10 dataset with an improvement of $1.3-1.7\times$ in GPU-hours and $1.5-1.7\times$ in the memory requirements during the architecture search stage. Moreover, the proposed frameworks do not require a well-engineered search space a priori, which is an important step towards fully automated design of DNN architectures.

% Differentiable neural architecture search is an efficient method to search for channel dimensions in DNNs. However, the existing solutions (i.e., differentiable channel masking) requires a well-designed search space, which hinders their practicality. In this work, we propose a novel technique that enables the search space to change based on the progress made during the search phase. The proposed technique allows searching for channel dimensions with loosely-defined initial conditions, improving the efficiency and practicality of channel dimension search. 

\keywords{Deep neural networks, neural architecture search, differentiable channel masking}
\end{abstract}

\input{sections/intro}

\input{sections/related_work}

\input{sections/background}

\input{sections/method}

\input{sections/results}

\input{sections/conclusion}

\bibliographystyle{splncs04}
\bibliography{refs}

\input{sections/appendix}

\end{document}

%% file: sections/intro.tex
\section{Introduction}

Deep neural networks (DNN) have become ubiquitous in numerous application domains such as computer vision and natural language processing. However, the performance and computational requirements of DNN models are highly dependent on their channel dimensions, namely a set of hyperparameters that define the number of filters in DNN layers. Finding the right set of channel dimensions for a DNN model plays a crucial role in achieving high performance under tight computational resource constraints, but it poses a challenging task for developers and engineers. Therefore, researchers have developed neural architecture search frameworks that automatically look for optimal channel dimensions of a DNN model\cite{Tan19CVPR, Wan20}.

The early versions of neural architecture search frameworks have resorted to reinforcement learning \cite{Pham18, Tan19CVPR, Zoph17, Zoph18}, evolutionary algorithms \cite{Marchisio20,Real19}, and Bayesian optimization \cite{Bergstra11} to search for optimal DNN architectures. Unfortunately, these methods have time and space complexities that increase combinatorially with the number of options that are defined in the search space, requiring excessive amounts of computational resources. To reduce the computational complexity of channel dimension search, Wan et al. \cite{Wan20} proposed the differentiable channel masking method, which mimics various channel dimensions by simply passing the feature maps through a set of binary masks. While the differentiable channel masking method significantly reduces the computational complexity of channel dimension search, it requires a search space that is carefully designed and tuned prior to the search phase, which hinders its practicality and usability in realistic scenarios.

In this work, we propose FlexCHarts, which utilizes a flexible search space that does not need to be defined a priori. We reformulate the problem of differentiable channel dimension search such that FlexCHarts does not only search for the optimal channel dimensions but also modifies the boundaries of the search space on-the-fly to add further flexibility and reach optimal channel dimensions. The proposed method relaxes the requirement of an a priori well-designed search space for channel dimension optimization and enables finding the optimal channel dimensions only with loosely-defined initial conditions of a search space, improving the practicality and usability of neural architecture search.

The rest of this paper is organized as follows: We first discuss the related work on neural architecture search, then we give a background information on the differentiable channel masking method, which forms the basis of our work. We then elaborate on the proposed FlexCHarts method as well as our novel dynamic channel allocation mechanism. Finally, we give the details of our experiments and discuss the results.

%% file: sections/related_work.tex
\section{Related Work}

Building DNN architectures that achieve high performance at low minimal hardware performance is a challenging task for developers and engineers. Thus, researchers have put considerable effort into developing effective neural architecture search methods to help automate the design of DNNs. Early work on neural architecture search adopted frameworks like reinforcement learning \cite{Pham18, Tan19CVPR, Zoph17, Zoph18}, evolutionary algorithms \cite{Marchisio20,Real19}, and Bayesian optimization \cite{Bergstra11}. Because these methods operate on a discrete search space and need to perform many trials while searching for an optimal architecture in an exponentially-increasing hyperparameter space, they require thousands of GPU-hours to find optimal DNN architectures, which greatly limits their applicability.  

To mitigate the prohibitive cost of architecture search, techniques such as weight-sharing \cite{Pham18} and one-shot search \cite{Bender18} have been proposed. While these techniques reduce the cost of each trial by allowing to reuse trained parameters, they still require many trials to find the optimal DNN architectures. To further improve the efficiency of neural architecture search, Liu et al. \cite{Liu19} proposed a differentiable neural architecture search, which employs a gradient descent optimizer to efficiently search for optimal DNN architectures. Furthermore, Wu et al. \cite{Wu19} proposed a hardware-aware differentiable neural architecture search, which enables to co-optimize both task accuracy and hardware metrics (e.g., latency). While these improvements drastically improved the computational efficiency of neural architecture search and the performance of the resulting DNN architectures, they did not address the challenges of searching for optimal channel dimensions, which is critical to the performance of DNNs.

\begin{figure}[t]
    \centering
    \includegraphics[width=0.8\linewidth, trim={5cm 0 0 0}, clip]{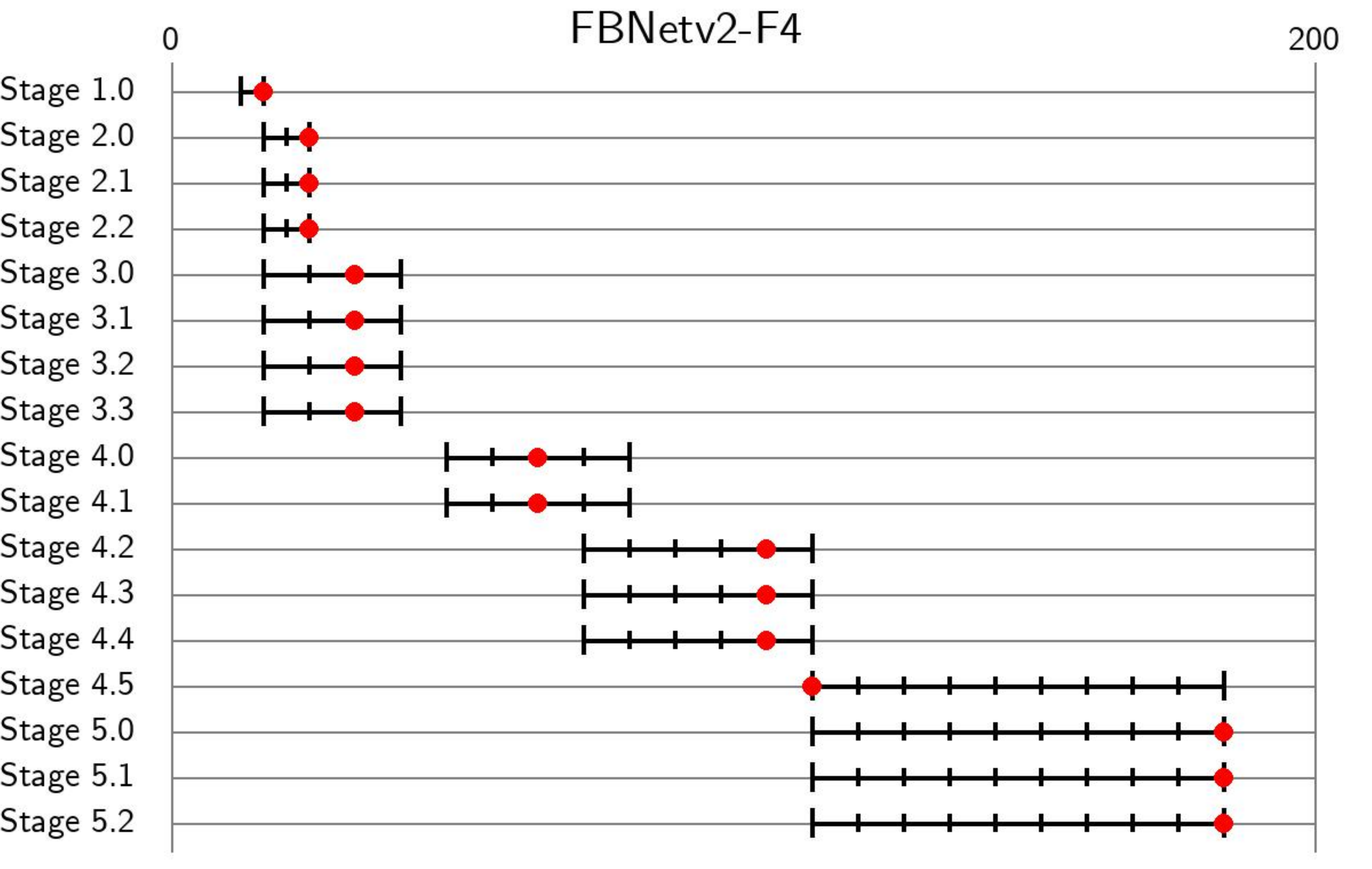}
    \caption{Prior work's search space for channel dimensions~\cite{Wan20}. Rows correspond to the channel range (between 0 and 200) of the layers in FBNetv2-F4 \cite{Wan20}. Ticks denote the options for channel dimensions and red circles represent the channel dimensions found. }
    \label{fig:range}
\end{figure}

To that end, Wan et al. \cite{Wan20} proposed differentiable channel masking method, which significantly improves the computational efficiency of neural architecture search with a search space that includes large numbers of options for channel dimensions. Unfortunately, the effectiveness of the proposed method is highly sensitive to the design of search space due to its requirements of a fixed search space. \autoref{fig:range} shows the search space of an existing differentiable neural architecture search framework for channel dimensions~\cite{Wan20}. We observe that the channel ranges in this search space are narrowed down to a few options out of a wide range using heuristics, while allowing only a limited degree of freedom for the optimizer. Unfortunately, determining a priori the channel range of each layer is nontrivial as it requires expert knowledge on DNN architectures. Also, the resulting search space is specific to the problem that it is designed for and it may not be transferable to the settings with different objectives and constraints. Moreover, when we investigate the channel dimensions found within this search space, as shown by red circles in \autoref{fig:range}, we observe that many channel dimensions are located at the boundary of the search space, which indicates that the optimal channel dimensions may lie outside of the engineered search space, resulting in a DNN architecture that does not correspond to the optimal solution. 

Few prior works have addressed the issues inherent to the fixed search spaces. Liu et al.~\cite{Liu18} proposed the \textit{progressive} neural architecture search, which gradually increases the complexity of the search space during optimization. Similarly, Ci et al.~\cite{Ci21} proposed the \textit{neural search space evolution} technique, which enables adding new operations to the search space as the architecture search progresses. Unfortunately, these methods are not applicable to the neural architecture search with channel dimension search. As a result, efficient search of channel dimensions without the restrictions of a fixed search space remains an open research question. Our work addresses these challenges and introduces a flexible search space for channel dimensions by proposing a novel differentiable neural architecture search to optimize for channel dimensions efficiently.

%% file: sections/background.tex
\section{Differentiable Channel Masking}

The input and output channel dimensions of a convolutional kernel have a significant impact on its performance and computational complexity. Increasing the number of channels typically leads to improved performance due to higher number of parameters at the expense of computational complexity, and vice versa. Thus, the channel dimensions must be carefully tuned while designing DNNs to obtain the required performance while attaining low computational complexity. With the number of kernels in modern DNNs exceeding hundreds, the task of finding the optimal channel dimensions for each kernel in a DNN is overwhelmingly time-consuming and costly. Therefore, researchers have developed automatic search methods for channel dimensions.

\begin{figure*}[t]
    \centering
    \includegraphics[width=1.\linewidth]{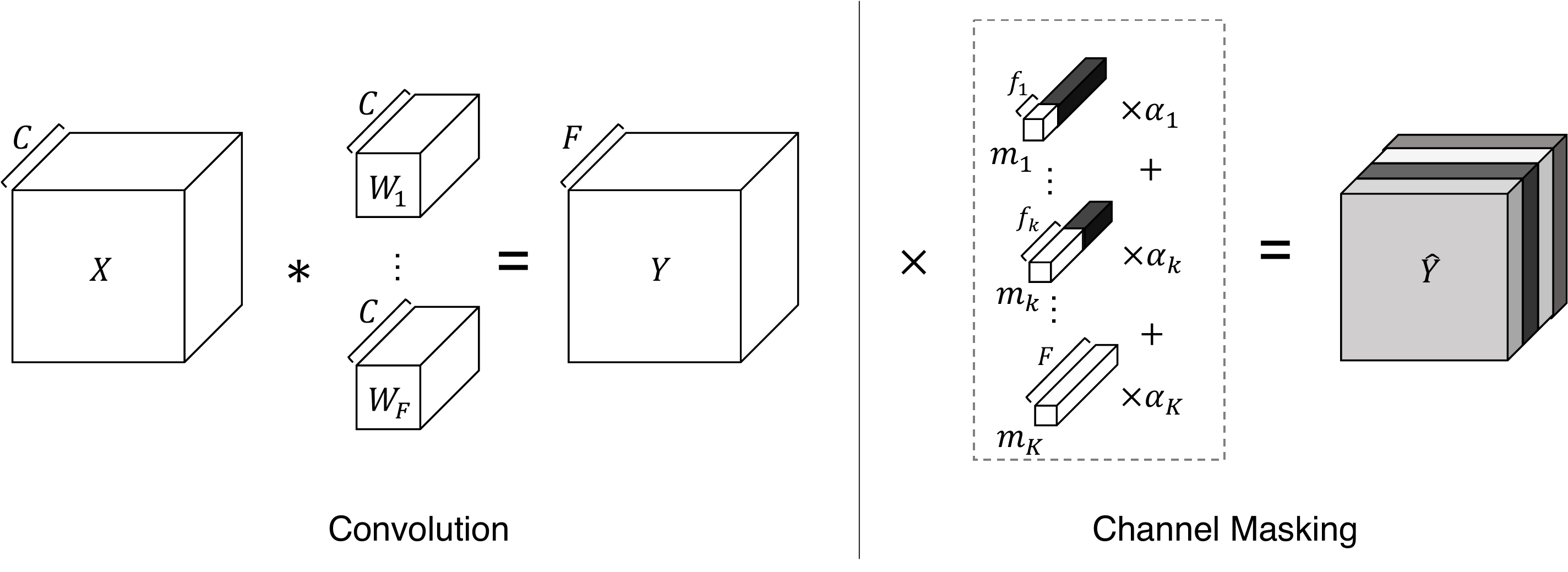}
    \caption{Illustration of a convolutional operation followed by channel masking to simulate various output channel dimensions.}
    \label{fig:channelmasking}
\end{figure*}

In standard search methods, each candidate channel dimension requires an additional convolutional kernel, which increases the computational cost and memory requirements of the search linearly with the number of channel options. The channel masking method, on the other hand, simulates various candidate channel dimensions on a single overparameterized kernel, incurring only minimal computational overhead. Because of its computational efficiency, we use the channel masking method as a basis for the proposed framework. In this section, we explain how the channel masking method works in more detail. 

Let us consider a convolutional kernel with a number of output channels $f$, which is to be chosen from a set of possible channel dimensions $S$, denoted as below:

\begin{equation}
    \label{eq:set}
    S = \{f_{k} \mid f_{k} \in \mathbb{Z}, k \in \mathbb{Z}, 1 \leq k \leq K \}
\end{equation}

\noindent where $f_{k}$ represents the possible channel dimensions and $K$ is the size of the set of possible channel dimensions. To evaluate the likelihood of $f_{k}$ being the optimal channel dimension for the given task, the channel masking method defines a set of trainable parameters $\{\alpha_{k} \mid \alpha_{k} \in \mathbb{R}, 1 \leq k \leq K \}$, where higher values of $\alpha_{k}$ indicate that $f_{k}$ is more likely to be the optimal channel dimension. 

Evaluating the impact of each $f_{k}$ on a DNN's overall performance and computational cost requires instantiating and training a separate kernel for each option, which increases the memory footprint and computational complexity of the search stage linearly with the number of options. To efficiently search for optimal channel dimensions with large numbers of options, the channel masking method instantiates and trains a single kernel and simulates various channel dimensions masking out a fraction of the channels in this kernel. 

\autoref{fig:channelmasking} illustrates how the channel masking method simulates various channel dimensions using a single kernel. Let $\mathbf{X}$, $\mathbf{W}$ and $\mathbf{Y}$ be the input activation, weight, and output activation of a convolutional kernel. $\mathbf{X}$ and $\mathbf{Y}$ have channel dimensions of $C$ and $F$,respectively. The channel masking method exploits the fact that any convolutional kernel with an output channel dimension $f_{k}$ that is smaller than $F$ can be obtained simply by selecting $f_{k}$ channels from $\mathbf{Y}$ and masking out the rest. For this purpose, the channel masking method instantiates a set of masks $\displaystyle \{m_{k} \mid k \in \mathbb{Z}, 1 \leq k \leq K \}$, where $m_k = (1)_{i=1}^{f_k} \cup (0)_{i=f_k+1}^{F}$. In other words, the first $f_k$ elements of $m_k$ are set to one whereas the remaining elements are zero; thus, the $m_k$ allows selecting the first $f_k$ channels of $Y$ and zeroes out the channels that are greater than $f_k$.

During the search phase, the DMaskingNAS method multiplies the output activation $\mathbf{Y}$ by the masks $m_k$ and calculate $\mathbf{\hat{Y}}$, which is the weighted sum of the output of simulated layers with various channel dimension using the following formula:

\begin{equation}
    \label{eq:weightedsum}
    \hat{Y}= \sum_{k=1}^{K}{g_{\tau}(\alpha_{k})m_{k}Y}
\end{equation}

\noindent where $g_{\tau}$ is a Gumbel softmax function with the temperature constant $\tau$ that maps the $\alpha$ values between 0 and 1~\cite{Wan20}. The expression in \autoref{eq:weightedsum} can be simplified by taking $\mathbf{Y}$ out of the summation, reducing the overhead of masking to only a weighted sum of low-dimensional masks. Therefore, the channel masking method simulates multiple channel dimensions with negligible computational overhead.

During the search phase, the channel masking method updates the $\alpha$ values with a gradient descent optimizer by minimizing the following loss function:

\begin{equation}
    \label{eq:loss}
    \min_{\alpha}\min_{W} \mathcal{L}_{acc}(\mathcal{N}_{\alpha, W}(x), y) + \lambda\mathcal{L}_{latency}(\mathcal{N}_{\alpha, W})
\end{equation}

where $\mathcal{N}_{\alpha, W}$ represents the network, and $x$ and $y$ represent training samples and ground-truth, and $\mathcal{L}_{acc}$ and $\mathcal{L}_{latency}$ represent the loss functions for classification accuracy and latency, respectively. The coefficient $\lambda$ controls the trade-off between accuracy and latency. As suggested by the prior work~\cite{Liu19}, the loss function is minimized by calculating the gradients using a first-order approximation in order to reduce the computational cost of the search. At the end of the search phase, the final channel dimensions are selected as the channel dimensions that correspond to the maximum $\alpha_{i}$, where $\displaystyle i=\underset{k}{\arg\max} \alpha_{k}$. 

%TODO: Add this to the paper version: $\mathcal{L}_{acc}$ is typically defined as a cross-entropy for classification tasks and $\mathcal{L}_{latency}$ is often calculated as the number of operations required to perform a forward pass, which is differentiable in terms of channel dimensions. 

The channel masking method permits to search for channel dimensions among various options with minimal computational overhead. However, the standard channel masking methods proposed in prior work can search only within a fixed range of channel dimensions~\cite{Wan20}, which hinders its effectiveness and practicality. Thus, in the next section, we introduce FlexCHarts, which is a differentiable channel masking method that searches for channel dimensions in a flexible range.

% Efficiency of the proposed method, this is why we build on top of this formulation
% The channel masking method allows simulating multiple channel dimensions with negligible overhead. The expression in \autoref{eq:weightedsum} can be simplified by taking $Y$ out of the summation, reducing the overhead of masking to only a weighted sum of low-dimensional masks, which is insignificant compared to the preceding four-dimensional convolutional operation. Therefore, in our proposed method that is explained in the following section, we adopt a similar approach and use a differentiable channel masking method.

%% file: sections/method.tex
\section{FlexCHarts}

To enable searching for channel dimensions in a flexible range, we first introduce the flexible channel masking method, which redefines the $\alpha$ variables so as to permit to change the range of channel dimensions as the search progresses. Then, we elaborate on our dynamic channel allocation mechanism, which modifies the kernels to accommodate for the changes in the channel dimension range during the search.

\begin{figure}[t]
    \centering
    \subfloat[Vanilla channel masking.\label{fig:dmaskalphas}]{%
      \includegraphics[width=0.7\textwidth]{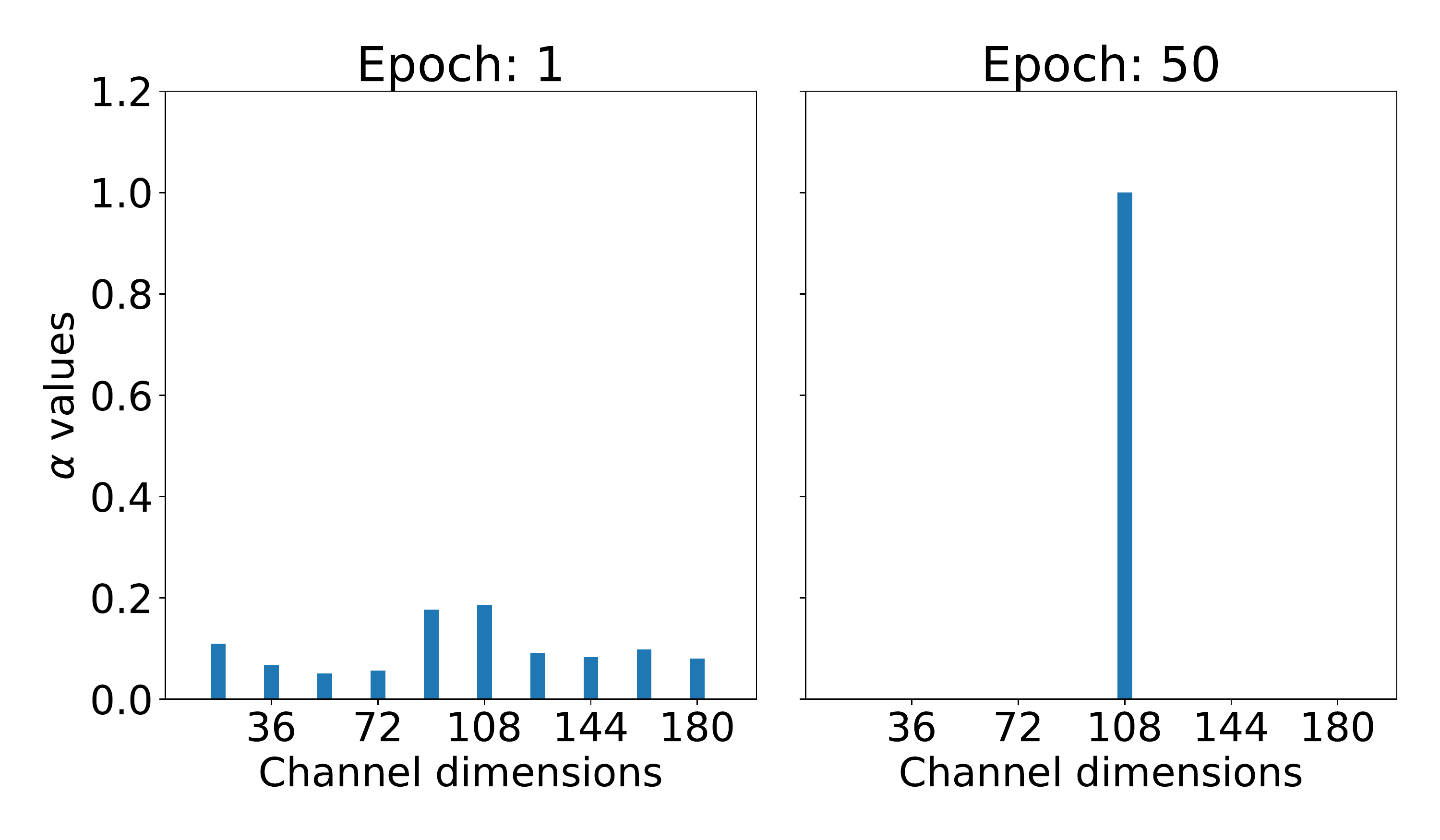}
    }
    \hfill
    \subfloat[FlexCHarts.\label{fig:flexchartsalphas}]{%
      \includegraphics[width=0.7\textwidth]{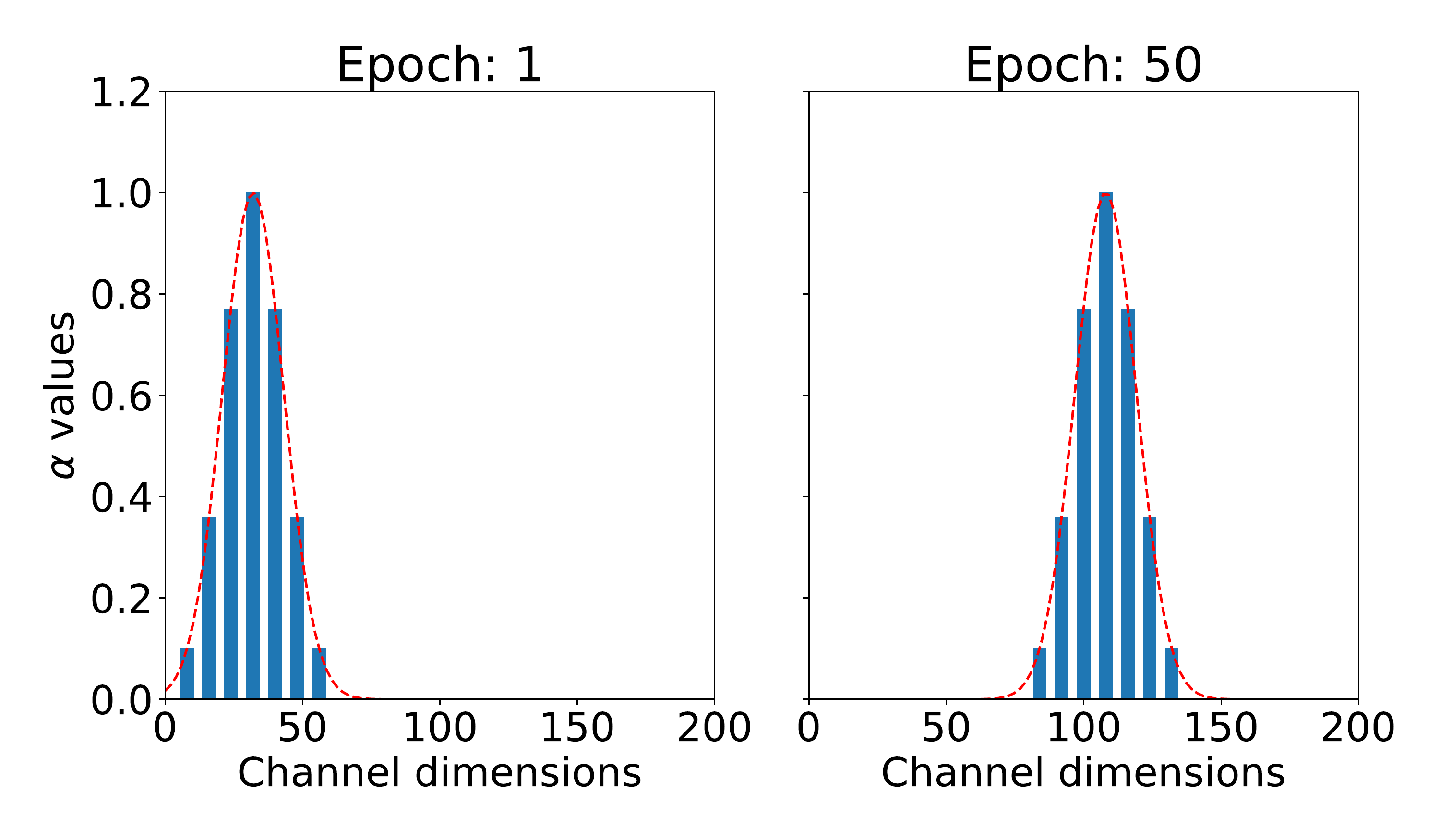}
    }
    
    \caption{An example of $\alpha$ values in vanilla channel masking method versus the proposed FlexCHarts methods between the first and last epoch of a search.}
    \label{fig:alphas}
\end{figure}

\subsection{Flexible channel masking}

FlexCHarts redefines the $\alpha$ variables in such a way that the channel dimension range can be changed on-the-fly during the search while still benefiting from the computational efficiency of the vanilla channel masking method. Instead of defining each $\alpha_k$ as an independent variables \cite{Wan20}, we define $\alpha_k$ as a smooth function of the channel dimension that it corresponds to, where it has the highest value at the center and close to zero at the edges of the range. While various smooth functions would be equally applicable, without loss of generality, we opt for the normal distribution to define $\alpha_{k}$ in this work due to its simplicity: 

\begin{equation}
    \label{eq:smoothfunction}
    \alpha_{k} = \exp(-\frac{1}{2}(\frac{f_k-\mu}{\sigma})^2)
\end{equation}

\noindent where $\mu$ and $\sigma$ are the mean and standard deviation of the exponential function, respectively. The formulation given above differs from the vanilla channel masking method in the sense that $\alpha$ is no longer a trainable variable. Instead, the proposed flexible channel masking method defines $\mu$ as the trainable variable and derives the $\alpha$ values from the exponential function given in \autoref{eq:smoothfunction}. This formulation smoothly adapts the $\alpha$ values to different channel ranges based on the information on the gradients of $\mu$. 

\autoref{fig:alphas} shows an example of how $\alpha$ values change in vanilla channel masking and in FlexCHarts, between the first and last epochs of a dimension search process. In vanilla channel masking method, as shown in \autoref{fig:dmaskalphas}, $\alpha$ values are initialized randomly. During the search, the $\alpha$ values are updated independently and eventually in the last epoch, the $\alpha$ value that corresponds to the optimal channel dimension becomes significantly higher than the others, setting the channel dimension for the final architecture. In contrast, the $\alpha$ values in FlexCHarts, as shown in \autoref{fig:flexchartsalphas}, are taken from the exponential function given in \autoref{eq:smoothfunction} (shown with red dashed line in the figure). The mean value of the exponential function is updated at every step of the search phase, gradually shifting the $\alpha$ values towards the optimal channel dimension.

The proposed reformulation of $\alpha$ variables in \autoref{eq:smoothfunction} has two main advantages. First, the given expression is already differentiable; thus, it eliminates the need for tuning additional optimization hyperparameters such as temperature and noise in Gumbel softmax to make the optimization amenable to solution with gradient descent as in the prior work~\cite{Wan20}. Second, thanks to $\alpha_k$ values that are close to zero at the edges of the range, shifting the range to larger or smaller values gently introduces the new parameters to the kernels, preventing abrupt changes in the loss value that would otherwise be detrimental to the search process. In short, the proposed formulation of $\alpha$ variables is more efficient, flexible and easier to tune than the vanilla channel masking method. Thus, it provides a better solution when searching for optimal channel dimensions in DNN architectures.

\subsection{Dynamic channel allocation}

In vanilla channel masking methods, the kernels do not require any dimension changes during the search phase as the channel range remains constant. However, with the reformulation of the $\alpha$ variables in FlexCHarts that permits to change the channel dimension range as the search progresses, the dimensions of the kernels must also be efficiently adapted during the search phase. We now elaborate on how we modify the kernels for dynamic channel dimension range.

Kernels in channel masking methods must have a number of channels equal to or greater than the maximum of the channel dimension range. Therefore, when the range of the channel dimensions shifts to higher values, we need to instantiate a larger kernel. Likewise, when the range of channel dimensions shifts to smaller values, a part of the kernel becomes redundant due to multiplication with an $\alpha_{k}$ value that is close to zero and is therefore no longer needed. As such, we can reduce the memory footprint and improve the computational efficiency of the search by switching to a smaller kernel. To adjust the dimensions of the kernels based on the changes in the channel dimension range, we introduce the dynamic channel allocation algorithm. 

There are two critical design considerations for the dynamic channel allocation algorithm. First, changing the kernel dimensions should have minimal impact on the on-going search to prevent loss of progress. Second, allocating new channels should incur only an insignificant computational overhead. To achieve the first condition, when the proposed dynamic channel allocation algorithm changes the dimensions of a kernel, it transfers the trained weights of the old kernel to the new one where applicable, which preserves the progress made in earlier training steps. For the latter objective, the proposed algorithm shall not react to changes in $\alpha_{k}$ values in every step. Instead, it waits until the end of an epoch to perform the changes to the kernels to reduce the computational overhead. Altogether, the proposed dynamic channel allocation algorithm enables changing the kernel dimensions with minimal impact on the search process and negligible computational overhead.

\begin{algorithm}[t]
    \SetKw{kwSupernet}{Supernet}
    \SetKw{kwOptimizer}{Optimizer}
    \SetKw{kwDataset}{Dataset}
    \kwSupernet{$\mathcal{N}_{\alpha,W}$ with parameters $W$ and $\alpha$\;}
    \kwOptimizer{weight optimizer $G_w$, arch optimizer $G_a$\;}
    \kwDataset{training: $D_{t}$, search: $D_{s}$\;}
    Initialize\;
    \For{all epoch $e$}{
        \For{all steps $s$}{
            Read training batch $b_t \leftarrow D_{t}$\;
            Backpropagate $\mathcal{N}_{\alpha,W}$ with $b_t$\;
            Update $W \leftarrow G_w$\;
            Read search batch $b_s \leftarrow D_{s}$\;
            Backpropagate $\mathcal{N}_{\alpha,W}$ with $b_s$\;
            Update $\alpha \leftarrow G_a$\;
        } 
        Update channel dimensions of $\mathcal{N}_{\alpha,W}$\;
    }
    \caption{FlexCHarts algorithm for channel dimension search with dynamic channel allocation.}
    \label{algo}
\end{algorithm}

In more details, the dynamic channel allocation proceeds as described in \autoref{algo}. The algorithm takes a Supernet $\mathcal{N}_{\alpha,W}$ as input, which consists of kernels with weights $\mathbf{W}$ and parameters $\alpha$, as well as the gradient descent optimizers $G_w$ and $G_a$ to update the weights and channel parameters, respectively. It also takes three datasets as inputs: equally sized $D_t$ and $D_s$ to train the weights and channel parameters, respectively. In each step of the algorithm, it reads a batch of samples from $D_t$, performs a backpropagation on the supernet, and updates the weights. Then, it repeats the same steps for channel parameters by reading a batch from $D_s$, performing a backpropagation, and updating them. When the same operations are performed for all samples in datasets $D_t$ and $D_s$, it updates the channel dimensions of the supernet based on the changes made to $\alpha$ values. We repeat the same operations for a predefined number of epochs.

In short, the proposed FlexCHarts algorithm permits to search for optimal channel dimensions in a flexible channel dimension range while automatically managing the changes in the supernet with minimal computational overhead.

%% file: sections/results.tex
\section{Experiments}

We now show the effectiveness of the proposed method through a number of experiments. In this section, we first give details about the search space, datasets, and hyperparameters that we use in the experiments, then we compare the proposed FlexCHarts method against the baseline methods through extensive experiments and discuss the results.

\subsection{Experimental setup}
\label{ch5:experimental-setup}

We perform experiments on a widely used image classification dataset, namely \cifar \cite{Krizhevsky2009} with a preprocessing pipeline for training that consists of a random crop of the input image with a size of $32$ and padding of $4$, random horizontal flip, normalization, and a cut-out with a length of 16~\cite{Devries17}. For the validation and test phases, we use only a normalization layer in the preprocessing pipeline. We use a batch size of $96$ for both the search and training phases. We perform all the experiments on an NVIDIA Tesla V100-SXM2 GPU with a 32GB memory.

For the search phase, we randomly split the training data set into two equally sized subsets to train the weights and channel parameters separately. We use a stochastic gradient descent (SGD) optimizer with a momentum coefficient of $0.9$, a weight decay of $3e-4$, a gradient clip of $5$ to train the weights. We initialize the learning rate of the SGD optimizer to $0.025$ and anneal it every step with a cosine annealing scheduler down to $0$ at the end of the last step. We use an Adam optimizer~\cite{Kingma14} with a learning rate of $0.1$, running average coefficients of $0.5$ and $0.999$, and a weight decay of $0$ to train the channel parameters. We use a dropout with a probability that starts at $0$ and linearly increases to $0.2$ until the end of the last step. The search phase takes 50 epochs to complete. After the search phase is completed, we train the DNN architecture with the discovered channel dimensions from scratch for 100 epochs to obtain its final accuracy. We also use an SGD optimizer in the training phase with the same hyperparameters as for the search phase.

As widely adopted by the community~\cite{Liu19, Wu19, Wan20}, we use a fixed stem and head stages at the beginning and end of our DNN architectures while we are searching for the optimal channel dimensions for the intermediate stages. The stem and head stages consist of convolutional blocks with kernel sizes of $3\times3$ and $1\times1$ and channel dimensions of 108 and 256, respectively. The intermediate block consists of 20 stages, each with a microarchitecture identical to the DARTS architecture~\cite{Liu19}. Each stage may have different channel dimensions as a result of the search phase while all the layers in a stage share the same channel dimension. For the training phase, we also use an auxiliary head that consists of three fully-connected layers with an auxiliary weight of 0.4 as proposed by the prior work~\cite{Liu19}.

\subsection{Performance of the differentiable channel search}
\label{sec:exp-flexcharts}
To evaluate the effectiveness of the proposed FlexCHarts method, we first compare it against the DMaskingNAS method, which has a fixed search space. Because the effectiveness and efficiency of DMaskingNAS method is highly sensitive to their predefined range of channel dimensions, we create two baselines that represent DMaskingNAS methods with small and large range of channel dimensions, which we simply refer to as DMask-small and DMask-large. The details of these search spaces are deferred to the Appendix.

To mimic target inference platforms with different resource constraints, we perform our experiments under low- and high-resource scenarios. For the low-resource scenario, we prioritize the computational requirements of the searched DNN architectures and target an inference latency under 0.3 millisecond per sample. In contrast, for the high-resource scenario, we prioritize the accuracy and aim for DNN architectures that achieve a top-1 test accuracy higher than 96\% on \cifar. We adjust the latency coefficients (i.e., $\lambda$ in \autoref{eq:loss}) to fulfill these accuracy and computational complexity requirements.

\begin{table}[t]
\caption{Results of the DMaskingNAS and FlexCHarts methods targeting low and high-resource scenarios. Check and cross marks indicate whether the requirement is satisfied or not.}
\begin{tabular}{c|c|c|c|c|c|c}
 \multicolumn{2}{c|}{} & \multicolumn{3}{c|}{Evaluation}  &  \multicolumn{2}{c}{Search}   \\ \hline

 \multirowcell{2}{Scenario} & Search        &  Top-1 acc.   &   Latency   &   FLOPS            & Search time   & Search  \\
                            &  algorithm    &  (\%)         &   (ms)      &  $\times 10^9$     & (GPU-hours)   & memory (GB)    \\ \hline
\multirowcell{3}{Low-resource  \\ ($<$0.3ms latency)} 
 & DMask-small              & 95.62   & 0.366 (\tikzxmark) &  0.258 & 2.63  & 10.3   \\
 & DMask-large              & 93.40   & 0.288 (\tikzcmark) &  0.095 & 5.75  & 28.3  \\
 & \textbf{FlexCHarts}      & 94.10   & 0.287 (\tikzcmark) &  0.093 & 3.36  & 16.6  \\  \hline
\multirowcell{3}{High-resource \\ ($>$96\% accuracy)}  
 & DMask-small              & 95.67 (\tikzxmark)   & 0.452 & 0.433 & 2.74  & 12.0 \\
 & DMask-large              & 96.06 (\tikzcmark)   & 0.606 & 0.736 & 5.78  & 28.3  \\
 & \textbf{FlexCHarts}      & 96.04 (\tikzcmark)   & 0.654 & 0.773 & 4.37  & 19.1
\end{tabular}

\label{tab:flexcharts-results}
\end{table}

\autoref{tab:flexcharts-results} summarizes the results of our experiments with the FlexCHarts, DMask-small, and DMask-large methods under the low- and high-resource scenarios. For the low-resource scenario, while the DMask-small method has the highest top-1 accuracy and lowest search time and memory, it fails to find a DNN architecture that achieves the target of 0.3 millisecond inference latency per batch due to its limited channel range. The DMask-large and FlexCHarts methods succeed to find DNN architectures that achieve the given inference latency target. However, the DMask-large method requires 5.75 GPU-hours and 28.3 GB of memory to find the DNN architecture as it needs to train a larger supernet whereas the FlexCHarts method finds an equivalent architecture only in 3.36 GPU-hours and using 16.6 GB of memory.

The FlexCHarts method also outperforms the DMaskingNAS method in the high-resource scenario. Due to its limited range of channel dimensions, DMask-small fails to find a DNN architecture that is large enough to achieve a top-1 accuracy greater than 96\%. In contrast, both DMask-large and FlexCHarts methods are able to find DNN architectures with the target accuracy requirements. Similarly to the low-resource scenario, the DMask-large method requires 5.78 GPU-hours and 28.3 GB of memory to achieve this goal whereas the FlexCHarts method finds an equivalent DNN architecture in 4.37 GPU-hours and using 19.1 GB of memory. 

These experiments clearly show that the proposed FlexCHarts method can find the channel dimensions that meet the requirements of varying resource constraints and optimization goals without the restrictions of a fixed search space. Moreover, it does not require to train a redundantly large supernet, thus it searches for the channel dimensions efficiently with lower GPU-hours and memory requirements than the DMaskingNAS methods. 

\subsection{Comparison with other dimension adaptation methods}

We now proceed with the experiments that compare the proposed FlexCHarts method against other channel dimension scaling methods. For this purpose, we use the following baselines: WideResnet architectures~\cite{Zagoruyko16}, which proposes to scale DNN architectures by simply multiplying its channel dimensions by a predetermined coefficient, and EfficientNet architectures~\cite{Tan19ICML}, which proposes compound scaling, in which the depth and width are scaled uniformly by a coefficient. 

In these experiments, we use WideResnet with a depth of 22 and width factors of $1$, $2$, and $4$ and EfficientNet with its four largest variants, namely \textit{B6}, \textit{B7}, \textit{B8}, and \textit{L2}, which requires a similar range of FLOPS with the DNN architectures found by FlexCHarts. To eliminate the performance discrepancies caused by differences in their implementations, we compare their computational complexity in terms of FLOPS. For different search runs with FlexCHarts, we used the latency coefficients ($\lambda$) varying between $1e-3$ and $1e-1$. We train all DNN architectures using the same training parameters given in \autoref{ch5:experimental-setup}. 

\begin{figure}[t]
    \centering
    \includegraphics[width=0.6\linewidth]{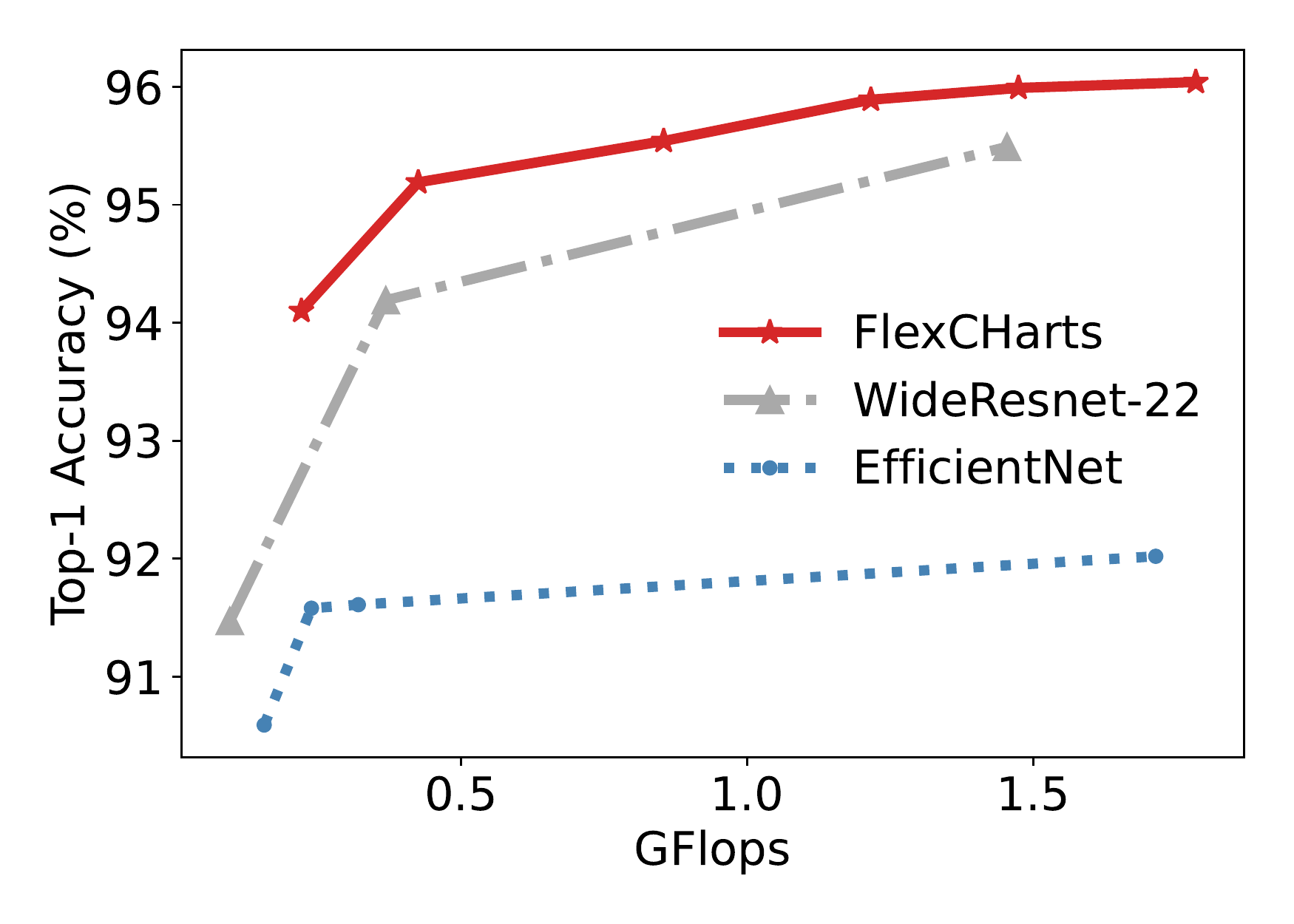}
    \caption{Accuracy on \cifar versus computational complexity in terms of FLOPS for various channel dimensions found by FlexCHarts as well as the baseline WideResnet and EfficientNet architectures.}
    \label{fig:results-cifar10}
\end{figure}

\autoref{fig:results-cifar10} shows the top-1 accuracy and FLOPs requirements of the DNN architectures found by the proposed FlexCHarts method as well as the baseline methods. Because FlexCHarts can automatically search the optimal channel dimensions within a flexible range, the architectures found by FlexCHarts achieve between 0.5-1\% and 2-4\% better accuracy than WideResnet and EfficientNet for similar FLOPS requirements, respectively. Moreover, the effectiveness of the channel scaling methods used in WideResnet and EfficientNet is highly sensitive to the design of initial DNN architectures, which requires heuristics and manual development efforts. In contrast, the proposed FlexCHarts method finds channel dimensions that achieve better accuracy and/or FLOPS and it does so in a completely automatic fashion.

%% file: sections/conclusion.tex
\section{Conclusion}

In this work, we addressed the limitations of neural architecture search methods that have fixed search space for channel dimensions. We redefined the architectural variables in the differentiable channel masking method to enable searching for channel dimensions in a flexible and smooth way. We also introduced a new dynamic channel allocation mechanism that allows changing the kernel dimensions efficiently during the search in order to dynamically adapt to the target channel dimensions. 

Through extensive experiments, we demonstrated that the proposed FlexCHarts framework finds optimal channel dimensions for DNN architectures under various resource constraints and performance objectives without the limitations of the existing methods that use fixed search spaces. Moreover, it searches the optimal channel dimensions faster and with reduced memory requirements than the existing methods. 

%% file: sections/appendix.tex
\appendix

\section{Appendix}
\label{ch:app-5}

\setlength{\tabcolsep}{10pt}
\renewcommand{\arraystretch}{1.}
\begin{table}[ht]
\centering
\caption{Channel ranges of DMask-small and Dmask-large baselines for the experiments in \autoref{sec:exp-flexcharts}.}
\begin{tabular}{c|ccc|ccc|ccc}
 & \multicolumn{3}{c|}{DMask-small} & \multicolumn{3}{c|}{DMask-large} & \multicolumn{3}{c}{DMask-systolic} \\ \cline{2-10} 
 Cell id & start & end & step & start & end & step & start & end & step \\ \hline
0 & 24 & 32 & 8 & 16 & 160 & 16 & 16 & 200 & 8\\
1 & 24 & 32 & 8 & 16 & 160 & 16 & 16 & 200 & 8\\
2 & 24 & 32 & 8 & 16 & 160 & 16 & 16 & 200 & 8\\
3 & 24 & 32 & 8 & 16 & 160 & 16 & 16 & 200 & 8\\
4 & 24 & 32 & 8 & 16 & 160 & 16 & 16 & 200 & 8\\
5 & 24 & 32 & 8 & 16 & 160 & 16 & 16 & 200 & 8\\
6 & 48 & 64 & 8 & 16 & 160 & 16 & 16 & 200 & 8\\
7 & 48 & 64 & 8 & 16 & 160 & 16 & 16 & 200 & 8\\
8 & 48 & 64 & 8 & 16 & 160 & 16 & 16 & 200 & 8\\
9 & 48 & 64 & 8 & 16 & 160 & 16 & 16 & 200 & 8\\
10 & 48 & 64 & 8 & 16 & 160 & 16 & 16 & 200 & 8\\
11 & 48 & 64 & 8 & 16 & 160 & 16 & 16 & 200 & 8\\
12 & 48 & 64 & 8 & 16 & 160 & 16 & 16 & 200 & 8\\
13 & 96 & 160 & 16 & 16 & 160 & 16 & 16 & 200 & 8\\
14 & 96 & 160 & 16 & 16 & 160 & 16 & 16 & 200 & 8\\
15 & 96 & 160 & 16 & 16 & 160 & 16 & 16 & 200 & 8\\
16 & 96 & 160 & 16 & 16 & 160 & 16 & 16 & 200 & 8\\
17 & 96 & 160 & 16 & 16 & 160 & 16 & 16 & 200 & 8\\
18 & 96 & 160 & 16 & 16 & 160 & 16 & 16 & 200 & 8\\
19 & 96 & 160 & 16 & 16 & 160 & 16 & 16 & 200 & 8\\
\end{tabular}

\label{tab:channel-range}
\end{table}

%% file: main.bbl
\begin{thebibliography}{10}
\providecommand{\url}[1]{\texttt{#1}}
\providecommand{\urlprefix}{URL }
\providecommand{\doi}[1]{https://doi.org/#1}

\bibitem{Bender18}
Bender, G., Kindermans, P., Zoph, B., Vasudevan, V., Le, Q.V.: Understanding
  and simplifying one-shot architecture search. In: Proceedings of the 35th
  International Conference on Machine Learning, {ICML}. vol.~80, pp. 549--558
  (2018)

\bibitem{Bergstra11}
Bergstra, J., Bardenet, R., Bengio, Y., K{\'{e}}gl, B.: Algorithms for
  hyper-parameter optimization. In: Advances in Neural Information Processing
  Systems 24: 25th Annual Conference on Neural Information Processing Systems.
  pp. 2546--2554 (2011)

\bibitem{Ci21}
Ci, Y., Lin, C., Sun, M., Chen, B., Zhang, H., Ouyang, W.: Evolving search
  space for neural architecture search. In: 2021 {IEEE/CVF} International
  Conference on Computer Vision, {ICCV} 2021, Montreal, QC, Canada, October
  10-17, 2021. pp. 6639--6649 (2021)

\bibitem{Devries17}
DeVries, T., Taylor, G.W.: Improved regularization of convolutional neural
  networks with cutout. arXiv (2017)

\bibitem{Kingma14}
Kingma, D.P., Ba, J.: Adam: {A} method for stochastic optimization. In: 3rd
  International Conference on Learning Representations, {ICLR} (2015)

\bibitem{Krizhevsky2009}
Krizhevsky, A.: Learning multiple layers of features from tiny images pp.
  32--33 (2009)

\bibitem{Liu18}
Liu, C., Zoph, B., Neumann, M., Shlens, J., Hua, W., Li, L., Fei{-}Fei, L.,
  Yuille, A.L., Huang, J., Murphy, K.: Progressive neural architecture search.
  In: Computer Vision - {ECCV} 2018 - 15th European Conference, Munich,
  Germany, September 8-14, 2018, Proceedings, Part {I}. vol. 11205, pp. 19--35
  (2018)

\bibitem{Liu19}
Liu, H., Simonyan, K., Yang, Y.: {DARTS:} differentiable architecture search.
  In: 7th International Conference on Learning Representations, {ICLR} 2019,
  New Orleans, LA, USA, May 6-9, 2019 (2019)

\bibitem{Marchisio20}
Marchisio, A., Massa, A., Mrazek, V., Bussolino, B., Martina, M., Shafique, M.:
  Nascaps: {A} framework for neural architecture search to optimize the
  accuracy and hardware efficiency of convolutional capsule networks. In:
  {IEEE/ACM} International Conference On Computer Aided Design, {ICCAD}. pp.
  114:1--114:9 (2020)

\bibitem{Pham18}
Pham, H., Guan, M.Y., Zoph, B., Le, Q.V., Dean, J.: Efficient neural
  architecture search via parameter sharing. In: Proceedings of the 35th
  International Conference on Machine Learning, {ICML} 2018,
  Stockholmsm{\"{a}}ssan, Stockholm, Sweden, July 10-15, 2018. vol.~80, pp.
  4092--4101 (2018)

\bibitem{Real19}
Real, E., Aggarwal, A., Huang, Y., Le, Q.V.: Regularized evolution for image
  classifier architecture search. In: The Thirty-Third {AAAI} Conference on
  Artificial Intelligence, {AAAI} 2019, The Thirty-First Innovative
  Applications of Artificial Intelligence Conference, {IAAI} 2019, The Ninth
  {AAAI} Symposium on Educational Advances in Artificial Intelligence, {EAAI}
  2019, Honolulu, Hawaii, USA, January 27 - February 1, 2019. pp. 4780--4789
  (2019)

\bibitem{Tan19CVPR}
Tan, M., Chen, B., Pang, R., Vasudevan, V., Sandler, M., Howard, A., Le, Q.V.:
  Mnas{N}et: Platform-aware neural architecture search for mobile. In: {IEEE}
  Conference on Computer Vision and Pattern Recognition, {CVPR} 2019, Long
  Beach, CA, USA, June 16-20, 2019. pp. 2820--2828. Computer Vision Foundation
  / {IEEE} (2019)

\bibitem{Tan19ICML}
Tan, M., Le, Q.V.: Efficient{N}et: Rethinking model scaling for convolutional
  neural networks. In: Proceedings of the 36th International Conference on
  Machine Learning, {ICML} 2019, 9-15 June 2019, Long Beach, California, {USA}.
  vol.~97, pp. 6105--6114 (2019)

\bibitem{Wan20}
Wan, A., Dai, X., Zhang, P., He, Z., Tian, Y., Xie, S., Wu, B., Yu, M., Xu, T.,
  Chen, K., Vajda, P., Gonzalez, J.E.: Fbnetv2: Differentiable neural
  architecture search for spatial and channel dimensions. In: 2020 {IEEE/CVF}
  Conference on Computer Vision and Pattern Recognition, {CVPR} 2020, Seattle,
  WA, USA, June 13-19, 2020. pp. 12962--12971 (2020)

\bibitem{Wu19}
Wu, B., Dai, X., Zhang, P., Wang, Y., Sun, F., Wu, Y., Tian, Y., Vajda, P.,
  Jia, Y., Keutzer, K.: Fbnet: Hardware-aware efficient convnet design via
  differentiable neural architecture search. In: {IEEE} Conference on Computer
  Vision and Pattern Recognition, {CVPR} 2019, Long Beach, CA, USA, June 16-20,
  2019. pp. 10734--10742 (2019)

\bibitem{Zagoruyko16}
Zagoruyko, S., Komodakis, N.: Wide residual networks. In: Proceedings of the
  British Machine Vision Conference 2016, {BMVC} (2016)

\bibitem{Zoph17}
Zoph, B., Le, Q.V.: Neural architecture search with reinforcement learning. In:
  5th International Conference on Learning Representations, {ICLR} 2017,
  Toulon, France, April 24-26, 2017, Conference Track Proceedings (2017)

\bibitem{Zoph18}
Zoph, B., Vasudevan, V., Shlens, J., Le, Q.V.: Learning transferable
  architectures for scalable image recognition. In: 2018 {IEEE} Conference on
  Computer Vision and Pattern Recognition, {CVPR} 2018, Salt Lake City, UT,
  USA, June 18-22, 2018. pp. 8697--8710 (2018)

\end{thebibliography}
